\documentclass{article} % For LaTeX2e
\usepackage{iclr2019_conference,times}

\usepackage{amsmath,amsfonts,bm}

\def\eqref#1{equation~\ref{#1}}
\def\1{\bm{1}}

\DeclareMathAlphabet{\mathsfit}{\encodingdefault}{\sfdefault}{m}{sl}
\SetMathAlphabet{\mathsfit}{bold}{\encodingdefault}{\sfdefault}{bx}{n}

\usepackage{hyperref}
\usepackage{url}
\usepackage{algpseudocode}
\usepackage{algorithm}
\newtheorem{theorem}{Proposition}

\usepackage[pdftex]{graphicx} 
\usepackage[caption=false,font=normalsize,labelfont=sf,textfont=sf]{subfig}
\usepackage{booktabs}
\newcommand{\qedwhitecustom}{\hfill \ensuremath{\Box}}

\title{Differentiable Greedy Networks}

\author{Thomas Powers$^{\dag}$\thanks{Work was done while Amanjit and Abdel-rahman  were at Amazon, Alexa AI and Thomas was as an intern at Amazon, Alexa AI.  \newline Corresponding authors: Thomas Powers and Rasool Fakoor },
Rasool Fakoor$^{\dag\dag}$,
Siamak Shakeri$^{\dag\dag}$,
Abhinav Sethy$^{\dag\dag}$,\\
\textbf{Amanjit Kainth}$^{\dag\ddag}$,
\textbf{Abdel-rahman Mohamed}$^{\ddag\ddag}$,
\textbf{Ruhi Sarikaya}$^{\dag\dag}$\\
$^{\dag}$University of Washington~~
$^{\dag\dag}$Amazon, Alexa AI~~
$^{\dag\ddag}$University of Toronto
$^{\ddag\ddag}$Facebook
}

\iclrfinalcopy % Uncomment for camera-ready version, but NOT for submission.
\begin{document}

\maketitle

\begin{abstract}
Optimal selection of a subset of items from a given set is a hard problem that requires combinatorial optimization. In this paper, we propose a subset selection algorithm that is trainable with gradient based methods yet achieves near optimal performance via submodular optimization. We focus on the task of identifying a relevant set of sentences for claim verification in the context of the FEVER task. Conventional methods for this task look at sentences on their individual merit and thus do not optimize the informativeness of sentences as a set. We show that our proposed method which builds on the idea of unfolding a greedy algorithm into a computational graph allows both interpretability and gradient based training. The proposed differentiable greedy network (DGN) outperforms discrete optimization algorithms as well as other baseline methods in terms of precision and recall.

\end{abstract}
%Even though  tasks such as fact extraction, fact verification, and question answering, we develop a new sentence retrieval method.
%Conventional evidence retrieval methods that look at lexical or semantic similarity between claim and sentence typically do not model dependencies between sentences. A submodular function can be used to capture dependencies between sentences, and despite making the sentence selection problem more difficult computationally, a near-optimal solution can be found efficiently using a simple forward greedy algorithm. Both of these approaches typically rely on pre-computed features, though, and can not make use of powerful supervised learning frameworks. The main contribution of this paper is a new network architecture derived from a submodular greedy optimization algorithm. By unfolding a greedy algorithm into a computational graph and making it differentiable, we can combine the advantages in interpretability and unsupervised initialization of conventional similarity ranking approaches with supervised learning techniques. The proposed differentiable greedy network (DGN) outperforms untrained discrete optimization algorithms and a conventional deep network in terms of recall and precision. 
\section{Introduction}\label{sec:intro}

In  this  paper,  we  develop  a  subset  selection  algorithm  that  is  differentiable  and  discrete,  which
can  be  trained  on  supervised  data  and  can  model  complex  dependencies  between  elements  in  a
straightforward and comprehensible way. This is of particular interest in  natural language processing tasks such as fact extraction,  fact verification, and question answering where the proposed optimization scheme can be used for evidence retrieval.

Conventional evidence retrieval methods that look at lexical or semantic similarity typically treat sentences or documents independently, potentially missing dependencies between them and therefore select redundant evidence. One way to address this shortcoming is by adding a diversity promoting submodular objective function~\citep{Tschiatschek2014ImageSummarizaiton, Lin2011Bilmes,lovasz1983submodular,fujishige2005submodular,sfotool}. Submodularity is a property of set functions that can be expressed by the notion of diminishing returns that allows near-optimal solutions to be found in polynomial time for NP-hard problems. 

A submodular set function is a function that maps sets to scalar values and has the property that the incremental value of the function computed with an additional element to an input set never increases as the input set grows. Submodular functions are defined by this natural diminishing returns property, which makes them well suited for tasks such as claim verification. With respect to a claim, the amount of relevant information in a set of sentences has diminishing returns as the set grows, meaning that the amount of additional information in an additional piece of evidence shrinks as the set of selected evidence grows. Thus, any relevancy-measuring function that is learned from data would potentially benefit
from a diminishing returns constraint as it would discount redundancy in favor of diverse but relevant evidence.  Claim verification often requires complicated induction from multiple sentences, so  promoting  diversity  among  selected  sentences  is  important  to  capture  all  facets  of  the  claim.
The resulting submodular optimization model can then handle dependencies between sentences and features, and despite making the sentence selection problem more difficult computationally, a near-optimal solution can be found efficiently using a simple forward greedy algorithm.

The  main  contribution  of  this  paper is a new optimization scheme which integrates continuous gradient-based and discrete submodular frameworks derived by  unfolding  a  greedy  optimization algorithm: the Differentiable Greedy Network (DGN). By unfolding a greedy algorithm into a computational graph, we can combine the advantages in interpretability and representation learning. We show that making a greedy algorithm differentiable and adding trainable parameters leads to promising improvements in recall@$k$ of 10\%-18\% and precision@$k$ of 5\%-27\% for a sentence selection task, where $k=1,3,5,7$ is the number of selected evidence sentences, on the Fact Extraction and Verification (FEVER) dataset~\citep{Thorne18Fever} and, with fewer parameters, performs very similarly to a conventional deep network. As the DGN is bootstrapping a greedy algorithm, it can be easily extended to work on other information retrieval tasks such as question answering as well as other problems that rely on greedy approaches. While more sophisticated neural architectures can deliver better performance, we focus on showing the power of our new optimization scheme on a simpler model.

In Section \ref{sec:relw}, we discuss related work in the domains of information retrieval, submodularity, and deep unfolding. In Section \ref{sec:model}, we define submodularity and present the proposed Differentiable Greedy Network (DGN). Section \ref{sec:exp} contains experiments and results for baseline models and DGN applied to sentence selection for the FEVER dataset as well as an ablation study. We draw conclusions in Section \ref{sec:conc}. Also, the attached Appendix~\ref{sec:app} contains an additional example demonstrating the utility of promoting diversity.

\section{Related Work}\label{sec:relw}
Evidence retrieval is a key part of claim verification or question answering which aims to provide reasoning and a measure of interpretability \citep{lei2016}. The FEVER dataset~\citep{Thorne18Fever} comes with a baseline evidence retrieval system, which returns the top $k$ sentences based on cosine similarity of term-frequency inverse-document-frequency (TF-IDF) features between the claim and candidate sentences. Models~\citep{VaswaniTransfoemer2017, huang2018fusionnet, Seo2016BidirectionalAF} that use attention on top of recurrent or convolutional neural networks can be adopted for this problem. However, all these types of models often focus only on learning similarities between claims and candidate evidence sentences, but neglect to model dependencies between the candidate sentences themselves, which is an advantage of a model that uses submodular functions which we focus on.

Central to our work here is the idea of deep unfolding \cite{hershey_deep_2014}  a technique that fills the continuum between the two extremes of generative models and deep learning models and combines the advantages of deep learning (discriminative training through backpropogation, ability to leverage large datasets) and generative models (ability to leverage domain knowledge, optimization bounds, fast training on smaller datasets).

Deep unfolding~\citep{hershey_deep_2014} has demonstrated a principled way to derive novel network architectures that are interpretable as inference algorithms by turning iterations of the inference algorithms into layers of a network. Previous work has unfolded Gaussian mixture models for multichannel source separation, the sequential iterative shrinkage and thresholding algorithm (SISTA) for sparse coding, and nonnegative matrix factorization (NMF) \citep{gregor_learning_2010,chen_end--end_2015,hershey_deep_2014,wisdom_interpretable_2016,wisdom_building_2017,le_roux_sparse_2015}. Particularly for sparse models, unfolding has only been applied to inference algorithms that use regularization approaches to promote sparsity. %On the discrete optimization side, algorithms typically use fixed or previously learned dictionaries \cite{cevher_greedy_2011,das_submodular_2011}.

%Deep unfolding has demonstrated a principled way to derive novel network architectures that are interpretable as statistical inference algorithms by turning iterations of the inference algorithms into layers of a network. Previous work has unfolded Gaussian mixture models for multichannel source separation, the sequential iterative shrinkage and thresholding algorithm (SISTA) for sparse coding, and NMF \citep{gregor_learning_2010,chen_end--end_2015,hershey_deep_2014,wisdom_interpretable_2016,wisdom_building_2017,le_roux_sparse_2015}. Particularly for sparse models, unfolding has only been applied to inference algorithms that use regularization approaches to promote sparsity. 

\cite{tschiatschek2018differentiable} considered subset selection through greedy maximization as well. Since output of their proposed method is differentiable as they can be interpreted as a distribution over sets, the proposed method can be combined with gradient based learning. However, their proposed method is different from us as it does stochastic selection algorithm which is not the case in our work. Moreover, we focus on sentence selection which comes with its own complexities. Others have proposed related versions of learning through discrete functions, including deep submodular functions (DSF) \citep{BilmesB17}, submodular scoring functions \citep{sipos2012}, iterative hard thresholding \citep{xin2016}, and the argmax function \citep{MenschB18}. The common thread of these related works is relaxation of the discrete functions. Our proposed network is distinct in that it learns the parameters of submodular function and encoder through the greedy optimization using a relaxed argmax function at training time. At test time, the model uses the argmax, and therefore retains it's submodular structure.

%{\huge ADD THE NIPS/IJCAI paper} 
\section{Differentiable Greedy Networks}\label{sec:model}
We start with a brief review of submodularity as it applies to our development of Differentiable Greedy Networks. The innovation of this paper stems from making greedy optimization of a submodular set function differentiable, but at test time, the performance guarantees are determined by the submodularity of the function, the constraints of the optimization problem, and the optimization algorithm used.

\subsection{Submodularity}
Submodularity is a property that describes set functions analogous to how convexity describes functions in a continuous space. Rather than exhaustively searching over all combinations of subsets, submodular functions provide a fast and tractable framework to compute a near optimal solution~\citep{lovasz1983submodular,fujishige2005submodular,sfotool}.

Let the set of available objects, known as the ground set, be denoted
as $V$. Submodularity can be expressed via the notion of diminishing returns, i.e., the incremental gain of the objective diminishes as the context grows. If we define the incremental gain of adding element $v$ to $A$ as $f \left( v | A \right) = f \left( A \cup \{ v \} \right) - f \left( A \right) $, 
then a submodular function $f$ is defined as satisfying
\begin{equation}
f \left( v | A \right) \geq f \left( v | B \right) \quad \forall A \subseteq B \subset V, \; v \notin B.
\label{eq:submod}
\end{equation}

%\begin{minipage}{\linewidth}[t]
\begin{figure}[t]
\centering
\vspace{-40pt}
\begin{minipage}{0.45\linewidth}
\begin{algorithm}[H]
\caption{$ \textsc{Forward Greedy} \left( V, k, f \right) $}
\label{alg:greedy}
\begin{algorithmic}[1]
\State $ \hat{A} \leftarrow \emptyset $
\While{ $ | \hat{A} | < k$ }
\State $ v^{\ast} \leftarrow \text{argmax}_{v \in V \backslash \hat{A}} f(v | \hat{A}) $
\State $ \hat{A} \leftarrow \hat{A} \cup \{ v^{\ast} \} $
\EndWhile\label{endwhile}
\State \textbf{return} $A$
%\vspace{53pt}
\end{algorithmic}
\end{algorithm}
\end{minipage}
\end{figure}
\vspace{-10pt}
%\end{minipage}

Submodular functions can be both maximized and minimized. Submodular functions that are maximized tend to exhibit concave behavior such as probabilistic coverage, sums of concave composed with monotone modular functions (SCMM), and facility location. Constrained monotone submodular function maximization algorithms include the forward greedy algorithm \citep{nemhauser_analysis_1978,fishermatroidbound} and continuous greedy algorithm \citep{vondrak08}. Likewise, constrained non-monotone submodular function maximization algorithms include local search algorithm \citep{Lee2009}, greedy-based algorithms \citep{gupta2010}, and other algorithms that use some combination of these \citep{mirzasoleiman2016fast}. In this paper, we focus on the forward greedy algorithm. 
Greedy optimization algorithms provide a natural framework for subset selection problems like retrieval, summarization, sparse approximation, feature selection, and dictionary selection. One can derive lower bounds if the objective function in a given subset selection problem is submodular \citep{nemhauser_analysis_1978}, and even for some non-submodular cases \citep{cevher_greedy_2011,das_submodular_2011,powers2016a}. 

For retrieval tasks like evidence extraction, submodular function optimization allows for dependence between potential pieces of evidence. For example, consider a problem where given a claim, the goal is to select sentences that help verify or refute the claim. A common simplifying assumption made in selection models is that the sentences are independent of each other, thereby ensuring the selection has linear time complexity in terms of the number of sentences. While this independence assumption doesn't necessarily adversely affect the relevancy of the selected sentences, it might lead to redundancy in the selected sentences.

The submodular function $f(\cdot)$ that we optimize is an SCMM given as follows:
\begin{equation}
\label{eq:scmm}
f_{\alpha} (A) = \sum_{u \in U}{ \alpha^{u} \; \text{log}( 1 + \sum_{a \in A}{h_a^{u}} )},
\end{equation}
where $\alpha^u \in \mathbb{R}_{+}, \, \forall u \in U$ are non-negative trainable parameters at index $u$ of the feature dimension that allow the functions to be tuned on data, and $h_a^u \in \mathbb{R}_{+}$ are the output features from the encoding layer for sentences $A$ and feature indices $U$. Specifically, they allow the network to identify the relative importance of different features $U$. For the sentence selection task that we test the DGN on in Section \ref{sec:exp}, these features correspond to the semantic similarities of a claim and potential evidence.

\begin{theorem}
\label{thm:scmm}
Let $m: 2^V \rightarrow \mathbb{R}_+$ be a modular function and $g: \mathbb{R} \rightarrow \mathbb{R}$ be a concave function. The function $f:  2^V \rightarrow \mathbb{R}$ defined as $f = g(m(A))$ is submodular.
\end{theorem}

\noindent \textbf{\textit{Proof:}} Given $A \subseteq B \subset V$ and $v \in V \backslash B$,  let $m(A) = x$, $m(B) = y$, and $m(\{v\}) = z$. Then $0 \leq x \leq y$, and $0 \leq c$. Since $g$ is concave, $g(x+z) - g(x) \geq g(y+z) - g(y)$, and therefore $g(m(A)+m(\{v\})) - g(m(A)) \geq g(m(B)+m(\{v\})) - g(m(B))$, which satisfies Equation (\ref{eq:submod}). \qedwhitecustom
 
Furthermore, submodularity is closed under nonnegative addition so Equation (\ref{eq:scmm}) is submodular if $\alpha^u, h_a^u \geq 0, \; \forall u \in U, a \in V $. 

Given the monotone submodular objective function, the optimization problem is to find the optimal subset $A^{\ast}$ that maximizes $f_{\alpha}( \cdot )$:
\begin{equation}
    A^{\ast} \in \underset{A \in V, |A| \leq k}{\text{argmax}} f_{\alpha} (A).
    \label{eq:argmax}
\end{equation}

Even though Problem (\ref{eq:argmax}) is NP-hard, we chose the objective function to be submodular, so a near optimal solution can be found in polynomial time \citep{nemhauser_analysis_1978}. Specifically, we can use the forward greedy algorithm, detailed in Algorithm \ref{alg:greedy}, to find a subset $\hat{A}$, such that
\begin{equation}
    f_{\alpha}(\hat{A}) \; \geq \; (1 - 1/e) \; f_{\alpha} ( A^{\ast} )
    \label{eq:approx}
\end{equation}
where $e$ is Euler's number, so $(1 - 1/e) \approx 0.63$. The guarantee in Equation (\ref{eq:approx}) ensures that any estimated solution set is no more than a constant factor worse than the true optimal set in terms of the objective function, and in practice the greedy algorithm often performs well above this tight lower bound.

\subsection{Differentiable Greedy Networks}
Now we present the Differentiable Greedy Network (DGN), an unfolded discrete optimization algorithm, which will allow for discriminative training of parameters while retaining the original guarantees of the forward greedy algorithm shown in Algorithm \ref{alg:greedy}.

The most important change to make the network differentiable is to the argmax function in line 3 of Algorithm \ref{alg:greedy}. During training, the argmax cannot be used directly because its gradients are not well-defined, so we approximate the argmax with softmax and use a temperature parameter $\tau$ to scale how close it approximates the argmax~\citep{Gumbel2017Jang}. This is an important parameter during training, as the temperature greatly affects the gradient flow through the network. As is discussed in Section \ref{sec:exp}, setting the temperature too low results in large gradients with high variance which makes training unstable and limits the gains in precision-recall. The temperature should not be too high either as the output starts to look more uniform and less like the argmax.

\begin{figure}[t]
\vspace{-23pt}
\begin{center}
\includegraphics[width=0.45\linewidth]{{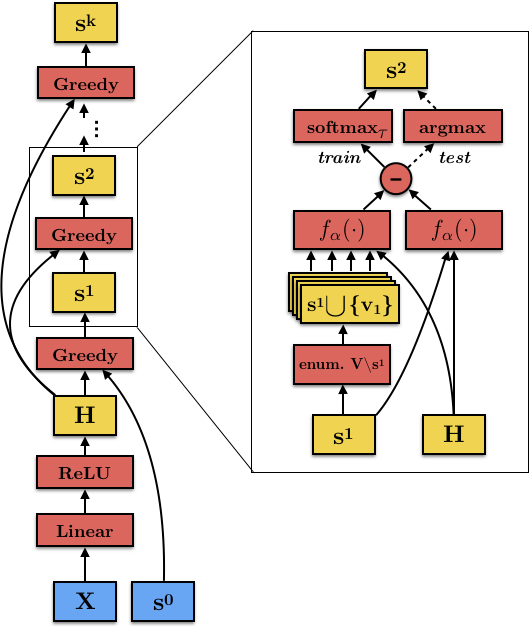}}
\end{center}
%\vspace{-12pt}
\caption{The Differentiable Greedy Network (DGN). The overall structure of the network on the left shows that the input features $X$ are run through a linear encoding layer and a ReLU activation function. The resulting non-negative features $H$ are then run through successive greedy layers along with the state vector which encodes the progression of sentence selections. The right diagram details a greedy layer. In the left branch, for greedy iteration $i$, the network enumerates the possible next states, $\{s^i \cup \{v\} : \forall v \in V \backslash s^i \}$. The network then evaluates these potential next states in the context of the current state $s^i$ and finds the sentence that provides the biggest increase in terms of $f_{\alpha}( \cdot )$, which is then added to the current state to form the new state $s^{i+1}$. During training, the selection is done through a softmax with temperature $\tau$ to differentiably approximate the argmax, and at test time we use the argmax to ensure the submodular guarantee of the model.}
\label{fig:dgn}
\vspace{-6pt}
\end{figure}

Figure \ref{fig:dgn} depicts the DGN as a computational graph. It is built out from Algorithm \ref{alg:greedy}, where lines 2-4 comprise the greedy layers. The overall structure of the network on the left shows that the input features, $X$, are run through a linear encoding layer follows by ReLU as an activation function. We choose the ReLU function specifically because the submodularity of the SCMM function requires non-negativity. The resulting non-negative features, $H$, are then run through successive greedy layers along with the state vector which encodes the progression of sentence selections. The right diagram details a greedy layer. In the left branch, for greedy iteration $i$, the network enumerates the possible next states, $\{\textbf{s}^i \cup \{v\} : \forall v \in V \backslash \textbf{s}^i \}$. The network then evaluates these potential next states in the context of the current state $\textbf{s}^i$ and finds the sentence that provides the biggest increase in terms of $f_{\alpha}( \cdot )$, which is then added to the current state to form the new state $\textbf{s}^{i+1}$. 

In some respects, the overall network looks conventional, especially the linear plus ReLU encoding layer. The connections between $H$ and the greedy layers are very reminiscent of a recurrent network where an input or hidden state is fed into layers with shared weights. The greedy layers, though, deviate significantly from conventional network architectures. Functionally, the greedy layers serve a similar purpose as an attention mechanism, but where attention would assign importance or relevance of a sentence to a claim or query, the submodular function in the greedy layer can be designed not only to measure the relevance of a sentence, but also explicitly model dependencies like redundancy between sentences and does so in a straightforward and interpretable manner. Examples of how the DGN models the sentence dependencies can be found in Section \ref{sec:exp}. The intuition behind how the DGN relates sentences together goes back to the submodularity of the objective function within the network. For the evidence extraction task described in Section \ref{sec:exp}, the input features are effectively similarity values between the claim and sentences across different feature indices. By applying a compressive concave function for a specific feature index $u$ aggregated across a set of sentences $A$ and summing over the feature indices $\forall u \in U$, a set of sentences that are similar to the claim but at different indices in the feature dimension would be preferred to a set of sentences that are equally similar but at the same indices.

\section{Experiments}\label{sec:exp}
We apply the DGN to the Fact Extraction and Verification (FEVER) dataset \citep{Thorne18Fever}. FEVER data consists of claims, classification of the claims as Supported, Refuted, or NotEnoughInfo (NEI), and evidence that the classification is based on. The claims and evidence come from Wikipedia. The baseline system for FEVER has two main steps: evidence retrieval and recognizing textual entailment (RTE). 
The dataset has $109,810$ verifiable claims in the training set, $13,332$ in the validation set and 6,666 in the test set. %A histogram of the number of labeled evidence sentences for the training set can be found in Figure \ref{fig:hist}. 
The baseline system selects $k$ sentences from a set of retrieved documents that have the highest TF-IDF cosine similarity to a claim using the DRQA toolkit~\citep{ChenDrQa17}. The FEVER scoring function computes the precision, recall, and F1 score for the retrieved sentences with respect to the labeled evidence sentences in verifiable claims. The evidence sentences are then run through the a baseline RTE network along with the claim to classify the claim as Supported, Refuted, or NEI. Notably, the best RTE baseline system drops from $88.00\%$ accuracy to $50.37\%$ when going from oracle evidence retrieval to the baseline retrieval system. Clearly, there is significant room to improve the evidence retrieval module. Also, the attached Appendix~\ref{sec:app} contains an additional example demonstrating the utility of promoting diversity.
%There are two RTE baselines, a multi-layer perceptron (MLP) with a single hidden layer and a decomposable attention (DA) model. The evidence sentences are then run through the MLP or DA network along with the claim to classify the claim as Supported, Refuted, or NEI. The simple MLP is rather effective for the entailment task, achieving an accuracy of 73.81\% when using oracle evidence. However, the accuracy drops dramatically to 40.63\% (and 19.42\% when excluding claims labeled NEI) when using the baseline evidence retrieval system. The decline extends to the DA model as well, dropping from 88.00\% accuracy to 50.37\% (and 23.53\% when excluding claims labeled NEI). Clearly, there is significant room to improve the evidence retrieval module. 
%\begin{figure}[t]
%\label{fig:hist}
%\begin{center}
%\includegraphics[width=0.7\linewidth]{{figures/feversenthist.png}}
%\end{center}
%\caption{Histogram of the number of evidence sentences in the FEVER %training set. Most have one or two sentences, but there is a heavy %tail.}
%\end{figure}

We narrow our focus to sentence retrieval given oracle documents. Most claims have one or two evidence sentences, but the distribution is fairly heavy-tailed. Based on the distribution, we set the max number of greedy iterations to $k=7$ in our experiments, which covers $93\%$ of the claims and provides a reasonable balance between coverage, precision, and computational complexity.
 %As expected, the recall grows as more sentences are selected but counterintuitively, using a smaller subset of pages to pull sentences from results in better evidence sentence recall. Using 5 pages has better performance across the board than using 20 or 25 pages. When selecting 20 or more sentences, using 10 pages This suggests that there are sentences in uninformative pages that have higher TF-IDF cosine similarity and are being chosen over sentences from the correct pages. Furthermore, this result suggests that cosine similarity over TF-IDF is not adequate to identify which sentences are informative, thereby motivating the use of alternative features, value functions, and optimization algorithms.
Claim verification often requires complicated induction from multiple sentences, so ideally, a more complicated model would be able to model the interactions between sentences. The next step is to introduce the concept of redundancy to the model by substituting in a diversity-promoting value function, SCMM, for the baseline value function, cosine similarity.

We use FastText word embeddings as input to the models~\citep{fasttext}. More specifically, we mean-pool the word embeddings of dimension $F$ for each claim and $D$ potential evidence sentences and element-wise multiply the claim feature vector with the sentences. Thus, the inputs $X \in \mathbb{R^{F \times D}}$ signify similarity scores between the claim and sentences for each of the feature dimensions. Connecting back to Equations (\ref{eq:scmm}) and (\ref{eq:argmax}), $F=|U|$ and $D=|V|$. Compared to TF-IDF features, the FastText word embeddings encode semantic structures of words that are more meaningful than simple lexical matching statistics. This lends itself well to the idea of promoting diversity. Consider a claim and three sentences, where the goal is to select the best two sentences as evidence. Using the SCMM in Equation (\ref{eq:scmm}), the forward greedy algorithm in Algorithm \ref{alg:greedy} would first select the sentence with the highest similarity to the sentence. If the other two sentences have are approximately equally similar to the claim, the SCMM will prefer the one that is similar across different feature dimensions than the chosen sentence. Therefore, this submodular framework will encourage semantic diversity among relevant sentences to a claim. 

The DGN encoder projects input features $X$ to a matrix $H \in \mathbb{R}_{+}^{F' \times D} $ by passing the features through two linear layers with ReLU non-linearity.  The state vector  $\textbf{s}$ is of dimension $D$. We perform random hyperparameter searches over learning rate, encoder output dimension $F'$, and softmax temperature $\tau$. The training loss is accumulated layer-wise cross entropy from output of the greedy layers(eq \ref{eq:CE}). We choose to compute the loss at each layer, rather than at the end to help better correct decisions made at the individual layers. We use Adam as the network optimizer with default parameters except for learning rate \citep{Adam}. The models are implemented in PyTorch \citep{pytorch}.

Let $\pmb{s}^{k} \in R^{D}$ denotes the state at layer k, which assigns a normalized score to each of the sentences. Labels are given by $L \subset \big\{1,...,D\big\}$. The Cross Entropy 
loss is computed as below:
% \begin{equation}\label{eq:CE}
%     Loss_{CE}(L,{\pmb{s}^{1},..,\pmb{s}^{k}}) = \sum_{j=1}^{min\big(K,|L|\big)} \sum_{i=1}^{D} \bigg[ \pmb{1}\{L_{j}=i\}log(\pmb{s}_{i}^{j}) + \pmb{1}\{L_{j} \neq i\}log(1-\pmb{s}_{i}^{j}) \bigg]
% \end{equation}
\begin{equation}\label{eq:CE}
    Loss_{CE}(L,{\pmb{s}^{1},..,\pmb{s}^{k}}) = \sum_{j=1}^{min(K,|L|)} \sum_{i=1}^{D} \bigg[ \mathbb{I}_{L_{j}=i} \cdot \text{log }\pmb{s}_{i}^{j} + \mathbb{I}_{L_{j} \neq i} \cdot \text{log}(1-\pmb{s}_{i}^{j}) \bigg]
\end{equation}

Where $\mathbb{I}_{A}$ is the indicator function for event $A$, and $K$ indicates the number of greedy layers($7$ in this paper).

\begin{table}[t]
\vspace{-35pt}
\caption{Evidence retrieval results. Each block of rows show recall, precision, and F1 when the models select $k = 7,5,3,1$ sentences as evidence. The DGN significantly outperforms the untrained baselines, When selecting fewer sentences, $k=5,3,1$, the recall improves recall while also producing increasing gains in precision. The DGN slightly, but consistently outperforms the Encoder.}
\label{tab:results}
\begin{center}
\begin{tabular}{l c | c c c }
Model & \ Recall@7 & Precision@7 & F1@7\\
\hline
Top-k       & 0.704 & 0.179 & 0.285\\
Greedy      & 0.706 & 0.179 & 0.286\\
Encoder     & 0.809 & 0.205 & 0.327\\
%DeepEncoder & 915K & 0.815 & 0.207 & 0.330\\
DGN         & \textbf{0.811} & \textbf{0.206} & \textbf{0.328}\\
%DDGN        & 4943K & 0.813 & 0.206 & 0.329\\
\hline
Model & \   Recall@5 & Precision@5 & F1@5 \\
\hline
Top-k       & 0.598 & 0.205 & 0.305 \\
Greedy      & 0.600 & 0.205 & 0.306\\
Encoder     & 0.734 & 0.251 & 0.374\\
%DeepEncoder & 915K & 0.741 & 0.254 & 0.378\\
DGN         & \textbf{0.738} & \textbf{0.253} & \textbf{0.377}\\
%DDGN        & 4943K & 0.741 & 0.254 & 0.378\\
\hline
Model & \   Recall@3 & Precision@3 & F1@3 \\
\hline
Top-k       & 0.438 & 0.244 & 0.313 \\
Greedy      & 0.439 & 0.245 & 0.314 \\
Encoder     & \textbf{0.615} & \textbf{0.342} & \textbf{0.440} \\
%DeepEncoder & 915K & 0.624 & 0.347 & 0.446 \\
DGN         & \textbf{0.615} & \textbf{0.342} & \textbf{0.440} \\
%DDGN        & 4943K & 0.619 & 0.345 & 0.443\\
\hline
Model & \   Recall@1 & Precision@1 & F1@1 \\
\hline
Top-k       & 0.194 & 0.322 & 0.242 \\
Greedy      & 0.194 & 0.322 & 0.242 \\
Encoder     & 0.353 & 0.588 & 0.441 \\
%DeepEncoder & 915K & 0.362 & 0.602 & 0.452 \\
DGN         & \textbf{0.356} & \textbf{0.593} & \textbf{0.445} \\
%DDGN        & 4943K & 0.370 & 0.615 & 0.462\\
\end{tabular}
\end{center}
\vspace{-15pt}
\end{table}

\begin{figure}[t]
\centering
\vspace{-30pt}
\includegraphics[width=0.45\linewidth]{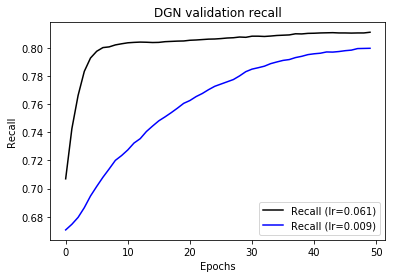}%
\vspace{-12pt}
\caption{Recall@7 on the validation set. The DGN seems to fit very quickly to the full set in terms of cross entropy, but the recall improves consistently. Decreasing the learning rate results in a smoother and consistently decreasing loss, but lower recall.}
\label{fig:recall}
\vspace{-4pt}
\end{figure}
%\begin{figure*}[t]
%\centering
%\subfloat[]{\includegraphics[width=0.48\linewidth]{{figures/dgn_loss.png}}%
%\label{fig:loss_full}}
%\hfil
%\subfloat[]{\includegraphics[width=0.48\linewidth]{figures/dgn_recall.png}%
%\label{fig:recall_full}}
%\caption{(a) shows training (solid lines) and validation (dashed lines) loss curves for the best performing model trained on the full dataset. (b) shows recall@7 on the validation set. The DGN seems to fit very quickly to the full set in terms of cross entropy, but the recall improves consistently. Decreasing the learning rate results in a smoother and consistently decreasing loss, but lower recall.}
%\label{fig:}
%\end{figure*}

The results in Table \ref{tab:results} show recall, precision and F1 scores for the tested models. The untrained baselines are a recreation of the FEVER system that selects the top $k$ sentences based on cosine similarity of sentence embeddings and a greedy algorithm that maximizes the submodular SCMM function in Equation (\ref{eq:scmm}). We also perform an ablation study to measure the effect of the greedy layers by removing them and measuring the performance of just the encoder (Encoder) and also test against a deep encoder model (DeepEncoder). The Encoder model has 172K trainable parameters. The DeepEncoder has an extra encoding layer compared to the Encoder having total of 915K parameters. Each block of rows of Table \ref{tab:results} show recall, precision, and F1 when the models select $k = 7,5,3,1$ sentences as evidence. The DGN outperforms the untrained baselines, improving recall@7 by 10\% while also improving precision. When selecting fewer sentences, $k=5,3,1$, the recall improves 14-18\% while producing increasing gains in precision 5\%, 10\%, and 27\%, respectively.

\begin{table}[t]
\caption{Evidence retrieval results for the DeepEncoder model and the DGN. Each row shows recall, precision, and F1 when the models select $k = 7$ sentences as evidence. The DGN achieves comparable performance of the DeepEncoder models with a fraction of the parameters in the best performing DeepEncoder.}
\label{tab:results2}
\begin{center}
\begin{tabular}{l c | c c c }
Model & \# Trainable Parameters & Recall@7 & Precision@7 & F1@7\\
\hline
DeepEncoder & \textbf{915K} & 0.815 & 0.207 & 0.330\\
DGN         & \textbf{167K} & 0.811 & 0.206 & 0.328\\
\end{tabular}
\end{center}
\vspace{-15pt}
\end{table}

While the performance of the Encoder and DeepEncoder as shown in Tables \ref{tab:results2} and \ref{tab:results} are very close to that of the DGN, the DGN maintains several advantages. First, the DGN is much more robust to class imbalances. In order to get the Encoder and DeepEncoder models to work, the positive samples (evidence sentences) needed to be weighted more heavily in the cross entropy loss. Moreover, the encoder models were very sensitive to this value--set too low the model rejected every sentence, and too high the model selects every sentence. The greedy algorithm regularizes the network to select $k$ sentences, though, and therefore is insensitive to the class imbalance. Second, since the $k$ greedy layers are performing $k$ iterations of submodular function optimization, the DGN explicitly and transparently models dependencies between functions given the right choice of submodular function. 

To illustrate how the dependencies encoded by the submodular function help, consider the following claim, which has two distinct facts that need to be verified:
%\begin{quote}
``Jarhead, a 2005 American biographical war drama, was directed by the award-winning auteur Sam Mendes.''
%\end{quote}
The first half of the claim asserts that Sam Mendes directed ``Jarhead,'' but looking at the labeled evidence the algorithm also needs to verify that he has won an award in the arts (presumably film). The DGN does just this in the first two greedy layers picking up on both facets of the claim with sentences $a$, $f_{\alpha}(\{a\}) = 14.48$, and $b$, $f_{\alpha}(\{b\}) = 12.86$:
%\begin{quote}
``He is best known for directing the drama film American Beauty (1999), which earned him the Academy Award and Golden Globe Award for Best Director, the crime film Road to Perdition (2002), and the James Bond films Skyfall (2012) and Spectre (2015),''
%\end{quote}
and 
%\begin{quote}
``Jarhead is a 2005 American biographical war drama film based on U.S. Marine Anthony Swofford's 2003 memoir of the same name, directed by Sam Mendes, starring Jake Gyllenhaal as Swofford with Jamie Foxx, Peter Sarsgaard and Chris Cooper.''
%\end{quote}
On the other hand, the DeepEncoder selects sentence $b$, verifying that Jarhead was directed by Sam Mendes,  but the next highest rated sentence is sentence $c$:
%\begin{quote}
``Samuel Alexander Mendes, (born 1 August 1965) is an English stage and film director.''
%\end{quote}
The DGN ranked both sentences $b$ and $c$ fairly highly, $f_{\alpha}(\{c\}) = 12.26$, but determined that sentence $c$  was more redundant with sentence $b$ than $a$ was, $f_{\alpha}(\{a,b\}) - f_{\alpha}(\{a\}) =  8.00$ compared to $f_{\alpha}(\{a,c\})  - f_{\alpha}(\{a\}) = 7.25$, so the score for $c$ dropped more than for $b$.

The DGN also had some interesting failure cases. A potential downside to promoting diversity in the selected sentences  is if the DGN selects the wrong sentence when the actual evidence sentences are ones that the submodular function deems redundant. For the claim ``Phoenix, Arizona is the capital of the Atlantic Ocean,'' which is easily refuted, the labelled evidence sentence $a$ is 
%\begin{quote}
``Phoenix is the capital and most populous city of the U.S. state of Arizona.''
%\end{quote} 
The DGN rates this sentence highly in the first greedy layer, $f_{\alpha}(\{a\}) = 13.23$, but not as highly as sentence $b$, $f_{\alpha}(\{b\}) = 15.99$:
%\begin{quote}
``In addition , Phoenix is the seat of Maricopa County and, at 517.9 square miles (1,341 km$^2$), it is the largest city in the state, more than twice the size of Tucson and one of the largest cities in the United States.''    
%\end{quote}
In the second layer, $f_{\alpha}(\{a,b\}) - f_{\alpha}(\{b\}) =  7.5$, which again lowers the score enough that the algorithm selects sentence $c$:
%\begin{quote}
``Despite this, its canal system led to a thriving farming community, many of the original crops remaining important parts of the Phoenix economy for decades, such as alfalfa, cotton, citrus, and hay (which was important for the cattle industry).''
%\end{quote}
The incremental gain of sentence $a$ continues to drop enough in subsequent layers to prevent selection as the DGN selects other sentence. Intuitively, this type of error seems especially problematic for claims that need to be refuted, specifically when the fallacy in the claim is seemingly unrelated to the evidence needed to refute it. There is very little connecting Arizona, for which Phoenix is the capital, and the Atlantic Ocean. In this case, we might prefer the algorithm to select more redundant sentences, because the extra information needed to refute it, Arizona, is not closely related to the correspondingly important part of the claim, Atlantic Ocean. 
%The oracle document retrieval did not include any pages about the Atlantic Ocean, but the wrongness inherent in asking about the capital of the Atlantic ocean adds to the difficulty.

Figure \ref{fig:recall} shows and recall on the validation set for the same two models. Both models fit very quickly to the full training and validation sets in terms of cross entropy, but the recall improves consistently. By lowering the learning rate, both the mean layer-wise cross entropy and recall smoothly improve, whereas raising the learning rate achieves a lower cross entropy after several epochs before increasing steadily as well as a higher and monotonically increasing recall. We believe this to be an artifact of the loss due to the greedy layers restriction of selecting a single sentence each. Looking back at Equation \ref{eq:CE}, at each outer sum we calculate the cross entropy assuming there is only one valid sentence. If there are more than one valid sentences, and the state, $\pmb{s}^{i}$, assigns large scores to those valid sentences, all except one are penalized by CE loss at that step. The ones penalized will be rewarded when calculating the loss at the later steps. This potentially can increase the loss when the model performance is improving. We observed the summed loss to be an acceptable trade-off as the goal is to maximize recall. Another advantage of the summed loss is that it reduces the issue of vanishing gradients as seen in recurrent networks. It will add direct gradient flow to the earlier layers of the unfolded network in the backward path.

Training the DGN was initially challenging. The choice of softmax temperature proved to be one of the most important hyperparameter settings. Our first thought was to make the temperature very low, $\tau = 0.01$, so that it was a close approximation of the argmax. This resulted in both vanishing and exploding gradients, because the loss was computing log of ones and zeros. To solve this, we experimented with two approaches. In the first approach, we set the temperature to be a number in the range (0.5,5) and is kept unchanged during the experiments. In the second approach, we apply temperature annealing, where in the beginning the temperature is set in the range as the first approach, and is annealed by 10\% every epoch. Our experiments showed that both approaches produce similar results. To make the training simpler, we chose the first approach. The experiments showed that temperatures in (3,6) range produce the best results.
\section{Conclusion}\label{sec:conc}
 In this paper, we have shown that unfolding a greedy algorithm into a computational graph, allowing us to retain the interpretability and unsupervised initialization of a conventional greedy sentence selection approach while benefiting from supervised learning techniques. The proposed differentiable greedy network (DGN) outperforms conventional discrete optimization algorithms in terms of both recall and precision. Furthermore, as sentence retrieval is often part of a larger pipeline as in the FEVER shared task, using a differentiable greedy network serves as a step towards an end-end trainable system.

\bibliography{iclr2019_conference}
\bibliographystyle{iclr2019_conference}

\section{Appendix}
\label{sec:app}
For an additional example of how submodularity is useful for diverse evidence extraction, consider the following claim:
\begin{quote}
``Stripes only featured women.''
\end{quote}
 For context, ``Stripes'' is a movie starring many male actors, so the claim is classified as Refuted. The DeepEncoder selects a true evidence $a$ sentence, 
 \begin{quote}``He first gained exposure on Saturday Night Live, a series of performances that earned him his first Emmy Award, and later starred in comedy films—including Meatballs (1979), Caddyshack (1980), Stripes (1981), Tootsie (1982), Ghostbusters (1984), Scrooged (1988), Ghostbusters II (1989), What About Bob? (1991), and Groundhog Day (1993),'' 
 \end{quote} 
and an incorrect one $b$ that is quite semantically similar, 
 \begin{quote}
``He also received Golden Globe nominations for his roles in Ghostbusters, Rushmore (1998), Hyde Park on Hudson (2012), St. Vincent (2014), and the HBO miniseries Olive Kitteridge (2014), for which he later won his second Primetime Emmy Award,''
 \end{quote}
both from actor Bill Murray's Wikipedia page. Initially, the DGN rates those two sentences highly as well, giving them the top two scores of $f_{\alpha}(\{a\}) = 7.87$ and $f_{\alpha}(\{b\}) = 7.21$, respectively. Therefore, the DGN selects the same first evidence sentence as the DeepEncoder, but in the second layer, the incremental value of $b$ drops significantly to $f_{\alpha}(\{a,b\}) - f_{\alpha}(\{a\}) =  4.74$, because it is highly redundant with $a$. Instead, the DGN selects $c$, \begin{quote}
``Stripes is a 1981 American buddy military comedy film directed by Ivan Reitman, starring Bill Murray, Harold Ramis, Warren Oates, P. J. Soles, Sean Young, and John Candy,''
\end{quote}
which also provides information refuting the claim in the form of a list of male actor names, but which is semantically different enough from $a$ to only drop from a score of $f_{\alpha}(\{c\}) = 6.73$ in the first greedy layer to $f_{\alpha}(\{a,c\})  - f_{\alpha}(\{a\}) = 5.21$ in the second greedy layer.

\end{document}